\documentclass[10pt,twocolumn,letterpaper]{article}

\usepackage[T1]{fontenc}
\usepackage[latin9]{inputenc}
\usepackage{color}
\usepackage{array}
\usepackage{multirow}
\usepackage{amsmath}
\usepackage{amssymb}
\usepackage{graphicx}
\usepackage{url}

\usepackage{mathptmx}

\usepackage{xspace}
\usepackage{color}
\usepackage{colortbl}

\usepackage{booktabs}                   % Publication quality tables

\definecolor{gray}{rgb}{0.35,0.35,0.35}
\definecolor{yellow}{rgb}{1,1,0.25}
\definecolor{MyBlue}{rgb}{0,0.2,0.8}
\definecolor{MyRed}{rgb}{0.8,0.2,0}
\definecolor{Red}{rgb}{1,0,0}
\definecolor{LightCyan}{rgb}{0.88,1,1}
\definecolor{MyGreen}{rgb}{0.0,0.5,0.1}
\definecolor{LightGray}{rgb}{0.9,0.9,0.9}

\def\red#1{\textcolor{red}{#1}}
\def\blue#1{\textcolor{blue}{#1}}

\newlength\paramargin
\newlength\figmargin
\newlength\secmargin

\usepackage{graphicx}
\graphicspath{{figures/}}

\usepackage{array}
\newcolumntype{L}[1]{>{\raggedright\let\newline\\\arraybackslash\hspace{0pt}}m{#1}}
\newcolumntype{C}[1]{>{\centering\let\newline\\\arraybackslash\hspace{0pt}}m{#1}}
\newcolumntype{R}[1]{>{\raggedleft\let\newline\\\arraybackslash\hspace{0pt}}m{#1}}

\long\def\ignorethis#1{}
\def\@onedot{\ifx\@let@token.\else.\null\fi\xspace}

\renewcommand{\paragraph}[1]{\noindent\textbf{#1}}
\usepackage{adjustbox}
\usepackage{caption}
\usepackage{enumitem}
\usepackage{float}
\usepackage{array}
\usepackage{multirow}

\usepackage[unicode=true,
 bookmarks=false,
 breaklinks=false,pdfborder={0 0 1},backref=page,colorlinks=true]
 {hyperref}

\makeatletter

 \usepackage{cvpr}
 \usepackage{times}
 \usepackage{amsmath}
 \usepackage{amssymb}
 \usepackage{colortbl}
 \usepackage{subcaption}
\newcommand{\algname}{DSRN }
\newcommand{\algnamens}{DSRN}

\newcommand*\samethanks[1][\value{footnote}]{\footnotemark[#1]}

\def\cvprPaperID{3225}
\cvprfinalcopy

\makeatother

\begin{document}

\title{Image Super-Resolution via Dual-State Recurrent Networks}

% \author{%
% \begin{tabular}{ccc}
% Wei Han* & Shiyu Chang* & Ding Liu  \tabularnewline
% Ins1 & Ins2\tabularnewline
% Address 1 & Address 2\tabularnewline
% \texttt{\small{}\href{mailto:firstauthor@i1.org}{firstauthor@i1.org}} & \texttt{\small{}\href{mailto:secondauthor@i2.org}{secondauthor@i2.org}}\tabularnewline
% \end{tabular}}
\author{Wei~Han$^1$\thanks{Authors contributed equally to this work},  Shiyu~Chang$^2$\samethanks,  Ding~Liu$^{1}$\thanks{Ding Liu and Thomas Huang's research works are supported
in part by US Army Research Office grant W911NF-15-1-0317.}, Mo~Yu$^2$, Michael~Witbrock$^2$,  Thomas S. Huang$^1$\samethanks\\
$^1$University of Illinois at Urbana-Champaign  $^2$IBM Research
\\
{\tt\small \{weihan3, dingliu2, t-huang1\}@illinois.edu}, \\
{\tt\small shiyu.chang@ibm.com}, \tt\small \{yum,witbrock\}@us.ibm.com}
\maketitle
\begin{abstract}
Advances in image super-resolution (SR) have recently benefited significantly from rapid developments in deep neural networks.  Inspired by these recent discoveries, we note that many state-of-the-art deep SR architectures can be reformulated as a single-state recurrent neural network (RNN) with finite unfoldings.   In this paper, we explore new structures for SR based on this compact RNN view, leading us to a dual-state design, the Dual-State Recurrent Network (\algnamens).  Compared to its single-state counterparts that operate at a fixed spatial resolution, \algname exploits both low-resolution (LR) and high-resolution (HR) signals jointly.   Recurrent signals are exchanged between these states in both directions (both LR to HR and HR to LR) via delayed feedback.  Extensive quantitative and qualitative evaluations on benchmark datasets and on a recent challenge demonstrate that the proposed \algname performs favorably against state-of-the-art algorithms in terms of both memory consumption and predictive accuracy. 
\end{abstract}

%%%%%%%%%%%%%%%%%%%%%%%%%%%%%%%%%%%%%%%%%%%%%%%%%%%%%%%%%%%%%%%%%%%%%%%%%%%%%%%
% Introduction
\section{Introduction}
In the problem of single-image super-resolution (SR), the aim is to recover a high-resolution (HR) image from a single low-resolution (LR) image.   In recent years, SR performance has been significantly improved due to rapid developments in deep neural networks (DNNs).  Specifically, convolutional neural networks (CNNs) and residual learning~\cite{he2016deep} have been widely applied in much recent SR work \cite{dong2014learning, fan2017balanced, kim2016accurate, kim2016deeply, lai2017deep, lim2017enhanced, tai2017image, tong2017image}.  

In these approaches, two principles have been consistently observed.   The first is that increasing the depth of a CNN model improves SR performance; a deeper model with more parameters can represent a more complex mapping from LR to HR images.  In addition, increasing network depth enlarges the size of receptive fields, providing more contextual information that can be exploited to reconstruct missing HR components.  The second principle is that adding residual connections (globally \cite{kim2016accurate}, locally \cite{kim2016deeply} or jointly \cite{tai2017image}) prevents the problems of vanishing and exploding gradients, facilitating the training of deep models.

While these recent models have demonstrated promising results, there are also drawbacks.  One major issue is that increasing the depth of models by adding new layers introduces more parameters,  and thus raises the likelihood of model overfitting.  At the same time, larger models demand more storage space, which is a hurdle to deployment in resource-constrained environments (\emph{e.g.} mobile systems).   To resolve this issue,  the Deep Recursive Residual Network (DRRN) \cite{tai2017image} inspired by the Deeply-Recursive Convolutional Network (DRCN) \cite{kim2016deeply} shares weights across different residual units and achieves state-of-the-art performance with a small number of parameters.

\begin{figure*}[t!]
\centering
\includegraphics[width=0.95\textwidth]{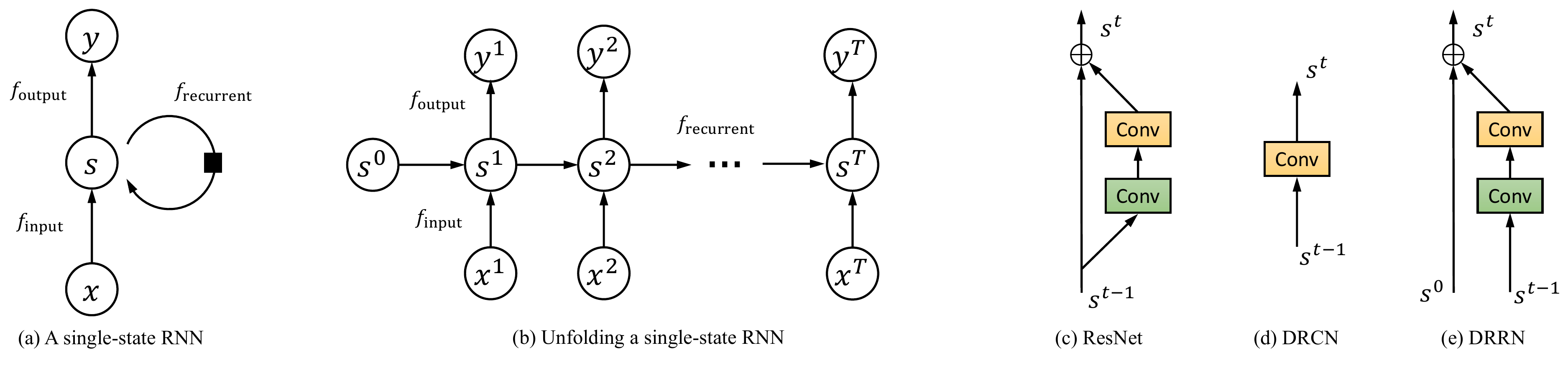}
\caption{(a) An example of a single-state RNN, which is characterized by an input state $x$, output state $y$ and a single recurrent state $s$.  The arrow links indicate the state transition function.  The black square represents the state transition function delayed for one time step.  (b) Finite unfolding ($T$ times) of a single-state RNN.  (c) - (e) The required recurrent function to make a single-state RNN equivalent to ResNet, DRCN, and DRRN, respectively.  Different colors of the ``Conv'' layers indicate different parameters. }
\label{fig:single-state-rnn}
\end{figure*}

Separate efforts \cite{chen2017dual, liao2016bridging, veit2016residual} in neural architectural design have recently shown that commonly-used deep structures can be represented more compactly using recurrent neural networks (RNNs).  Specifically, Liao and Poggio \cite{liao2016bridging} demonstrated that a weight-sharing Residual Neural Network (ResNet) \cite{he2016deep} is equivalent to a shallow RNN.   Inspired by their findings, we first explore the connections between the neural architectures of existing SR algorithms and their compact RNN formulations.   We note that previous SR models with recursive computation and weight sharing, including DRRN and DRCN,  work at a single spatial resolution (bicubic interpolation is first applied to upscale LR images to a desired spatial resolution).  This enables their model structures to be represented as a unified single-state RNN.  Thus, both DRRN and DRCN can be viewed as a finite unfolding in time of the same RNN structure, but with different transition functions.  This is illustrated in Figure~\ref{fig:single-state-rnn}, and will be discussed in detail in Section~\ref{sect:single-state}.  It is worth mentioning that we follow the terminology used in \cite{liao2016bridging}, where a ``state'' can be considered as corresponding to a ``layer'' in the normal RNN setting.

Based on this compact RNN view of state-of-the-art SR models,  in this paper we explore new structures to extend the frontier of SR.   The first approach in improving a conventional RNN model is generally to make it multi-layer.  We apply this experience in designing the SR architecture in our compact RNN view by adding an additional state, rendering our model a Dual-State Recurrent Network (\algnamens), where the two states operate at different spatial resolutions.  Specifically, the bottom state captures information at LR, while the top state operates in the HR regime.  As with a conventional two-layer stacked RNN, there is a  connection from the bottom to the top state via deconvolutional operations.  This provides information flow from LR to HR at every single unrolling time.   In addition, to allow information flow from previously predicted HR features to LR features, we incorporate a delayed feedback mechanism \cite{chung2015gated} from the top (HR) state to the bottom one.  The overall structure of the proposed \algname is shown in Figure~\ref{fig:dual-state-rnn}, which not only utilizes parameters efficiently but also allows both LR and HR signals to contribute jointly to learning the mappings.  

To demonstrate the effectiveness of the proposed method, we compare \algname with other recent image SR approaches on four common benchmarks \cite{bevilacqua2012low, huang2015single, martin2001database, zeyde2010single} as well as on the DIV2K dataset from the "New Trends in Image Restoration and Enhancement workshop and challenge on image super-resolution (NTIRE SR 2017)" \cite{agustsson2017ntire}. Extensive experimental results validate that \algname delivers higher parameter efficiency, low memory consumption and high restoration accuracy.

\section{Related Work}
Single image SR has been widely studied in the past few decades and has an extensive literature.
In recent years, due to the fast development of deep learning, significant progress has been made in this field.  Dong \emph{et al.} \cite{dong2014learning} first exploited a fully convolutional neural network, termed SRCNN, to predict the nonlinear LR-HR mapping.  It demonstrated superior performance to many other example-based learning paradigms, such as nearest neighbor \cite{freeman2002example}, sparse representation \cite{yang2010image,yang2012coupled}, neighborhood embedding \cite{chang2004super, timofte2014a+}, random forest \cite{schulter2015fast}, \emph{etc.}   
Although all layers of a SRCNN are trained jointly in an end-to-end fashion, conceptually the network is split into three stages: patch representation, non-linear mapping, and reconstruction.  

Much of the later work follows a similar network design with more complicated building blocks or advanced optimization techniques~\cite{shi2016real,liu2016robust,dong2016accelerating,liu2016learning}.  
Wang \emph{et al.}\ \cite{wang2015deep} proposed
a sparse coding network (SCN) that encodes a sparse representation prior for image SR and can be trained end-to-end, demonstrating the benefit of domain expertise in sparse coding for image SR. 
Both external and self examples were utilized to synthesize the HR prediction via a neural network in \cite{wang2015self}. 

Inspired by the success of very deep models \cite{he2016deep} on ImageNet challenges \cite{deng2009imagenet}, Kim \emph{et al.} \cite{kim2016accurate} proposed a very deep CNN, VDSR, which stacks 20 convolutional layers with $3 \times 3$ kernels.  Both residual learning and adjustable gradient clipping are used to prevent vanishing and exploding gradients.  However, as the model gets deeper, the number of parameters increases.  To control the size of the model, DRCN introduces 16 recursive layers, each with the same structure and shared parameters.  Moreover, DRCN makes use of skip connections and recursive supervision to mitigate the difficulty of training.  Tai \emph{et al.} \cite{tai2017image} discovered that many residual SR learning algorithms are based on either global residual learning or local residual learning, which are insufficient for very deep models.  Instead, they proposed the DRRN that applies both global and local learning while remaining parameter efficient via recursive learning.  More recently,  Tong \emph{et al.} \cite{tong2017image} proposed making use of Densely Connected Networks (DenseNet) \cite{huang2016densely} instead of ResNet as the building block for image SR.  They demonstrated that the DenseNet structure is better at combining features at different levels, which boosts SR performance.  

Apart from deep models working on bicubic upscaled input images, Shi \emph{et al.} \cite{shi2016real} used a compact network model to conduct convolutions on LR images directly and learned  upscaling filters in the last layer, which considerably reduces the computation cost.  Similarly, Dong \emph{et al.} \cite{dong2016accelerating} adopted deconvolution layers to accelerate SRCNN in combination with smaller filter sizes and more convolution layers. 
However, these networks are relatively small and have difficulty capturing complicated mappings owing to limited network capacity.
The Laplacian Pyramid Super-Resolution Network (LapSRN) \cite{lai2017deep} works on LR images directly and progressively predicts sub-band residuals on various scales.  
Lim \emph{et al.} \cite{lim2017enhanced} proposed the Enhanced Deep Super-Resolution (EDSR) network and a multi-scale variant, which learns different scaled mapping functions in parallel via weight sharing.  

It is noteworthy that most SR algorithms minimize the mean squared reconstruction error (\emph{i.e.} via $\ell_2$ loss).   They often suffer from regression-to-the-mean due to the ill-posed nature of single image SR, resulting in blurry predictions and poor subjective scores.  To overcome this drawback, Generative Adversarial Networks have been used along with perceptual loss for SR \cite{ledig2016photo,sajjadi2017enhancenet}. Subjective evaluation by mean-opinion-score showed huge improvement over other regression-based methods.

Our work is also strongly related to and built upon the idea of viewing a ResNet as an unrolled RNN.  It was first proposed in \cite{liao2016bridging}, which aids understanding of a family of deep structures from the perspective of RNNs.   Later, Chen et al. \cite{yunpeng2017sharing} unified several different residual functions to provide a better understanding of the design of DNNs with high learning capacity.  Recently,  the equivalence to RNNs has been further extended to DenseNet.  Based on this finding, Dual Path Networks \cite{chen2017dual} were proposed and showed superior performance to DenseNet and ResNet in a varity of applications. 

\section{Single-State Recurrent Networks}
\label{sect:single-state}
In this section, we first revisit the discovery that a ResNet with shared weights can be reformulated as a recurrent system.   Then, based on this view, we unite the recent development of SR models with such RNN reformulations to show DRCN and DRRN are structurally equivalent to an unrolled single-state RNN. 

To establish the equivalence, we adopt the commonly used definition of a RNN, which is characterized by a set of states and transition functions among the states.  A RNN often consists of the input state, output state, and the recurrent states.  Depending on the number recurrent states, we describe RNNs as ``single-state'' (\emph{i.e.} one recurrent state) or ``dual-state'' (\emph{i.e.} two recurrent states).  An illustration of a single-state RNN is shown in Figure~\ref{fig:single-state-rnn}(a).  The input, output, and recurrent states are represented as $x$, $y$ and $s$ respectively.  The arrow link indicates the state transition function.  The square on the directed cycle indicates that the recurrent function travels one time step forward during the unfolding.  Interested readers are referred to \cite{zhang2016architectural} for detailed information on this general formulation of a RNN.  

Based on Figure \ref{fig:single-state-rnn}(a), we unfold along the temporal direction to a fixed length $T$.  The unfolded graph is shown in figure \ref{fig:single-state-rnn}(b), and the dynamics of a single-state RNN can be characterized by:
\begin{equation}
% \small
\begin{aligned}
s^t & =f_{\text{input}}(x^t)+f_{\text{recurrent}}(s^{t-1}) \\
y^t &=f_{\text{output}}(s^t),
\end{aligned}
\end{equation}
where the upper script $t$ indicates the $t$-th unrolling.  The parameters of $f_{\text{input}}$, $f_{\text{output}}$, and $f_{\text{recurrent}}$ are often time-independent, which means these parameters are reused at every unfolding step.  This allows us to unify ResNet, DRCN, and DRRN as unrolled networks with the same recurrent structure but with the different realizations of $f_{\text{recurrent}}$ and different rules of parameter sharing. 

\vspace*{0.05in}
\noindent {\bf ResNet}:  We consider a ResNet in its simplest form without any down-sampling or up-sampling operations.  In other words, both of the spatial dimensions and feature dimensions remain the same across all intermediate layers.  To render Figure~\ref{fig:single-state-rnn}(b) equivalent to a ResNet with $T$ residual blocks, one possible technique is to make:
\begin{itemize}
\item $s^0$ be the input image $I$ or a function of $I$.
\item $x^t = 0, \forall t \in \{1, \dots T\}$, and $f_{\text{input}}(0) = 0$.  Thus, the state transition becomes $s^t = f_{\text{recurrent}}(s^{t-1})$.
\item The recurrent function $f_{\text{recurrent}}$ be the same as a conventional residual block, which contains two convolutional layers with skip connections as shown in Figure 
~\ref{fig:single-state-rnn}(c). Differences in color indicate different sets of parameters. 
\item The prediction state $y^t$ be calculated only at the time $T$ as the final output.
\end{itemize}
It is worth mentioning that the only difference between an unrolled RNN following the above definitions and a conventional ResNet is that the parameters in $f_{\text{recurrent}}$ need to be reused among all residual blocks.

\vspace*{0.05in}
\noindent {\bf DRCN}: To realize the DRCN expressible by the same single-state RNN, we define $s^0$ and $x^t$ in the same way as for the ResNet.  Since DRCN recursively applies only a single convolutional layer to the input feature map 16 times, with the parameters of the layer reused across the whole network, we could use a single convolutional layer to express $f_{\text{recurrent}}$.  The graph is illustrated in Figure~\ref{fig:single-state-rnn}(d).  Moreover, unlike the ResNet where the output is predicted only at the end of unfolding, DRCN utilizes recursive supervision, which generates an output $y^t$ at every unfolding $t$.  
The final HR prediction of DRCN is the weighted sum of the outputs at every unfolding $t$.
%Both predicting at the end of the sequence or every time step commonly exist in the field of natural language processing \cite{xing2010brief}, which is often referred as sequence classification and sequence labeling, respectively.  

\begin{figure*}[t!]
\begin{centering}
\includegraphics[width=0.9\textwidth]{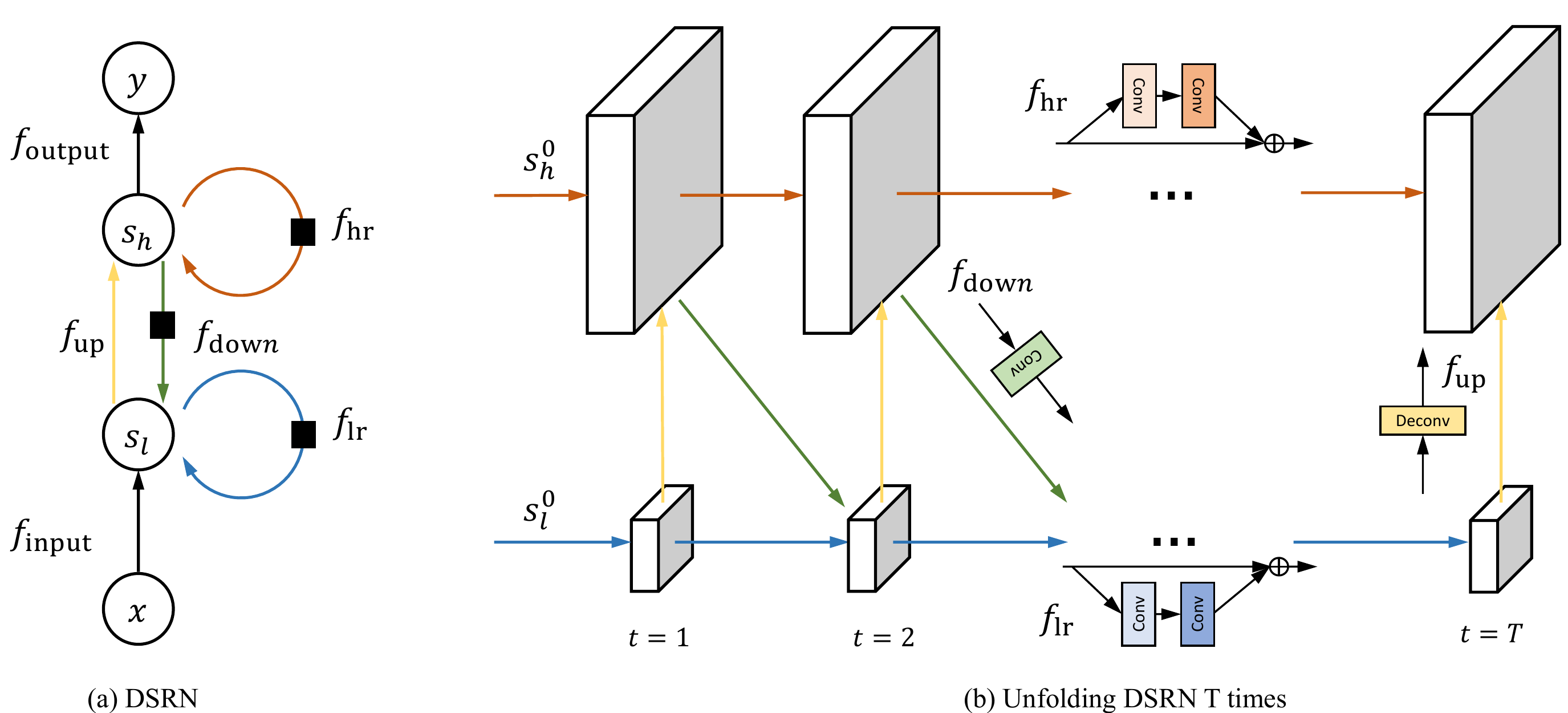}
\par\end{centering}
\vspace*{0.05in}
\caption{(a) The recurrent representation of the proposed \algnamens, whose graph definition is the same as Figure \ref{fig:single-state-rnn}(a). (b) The unrolled \algnamens. Edges with the same color have identical state transition functions and shared parameters.  The structures of four specific transition functions have been illustrated correspondingly.  ``Conv'' blocks with different colors indicate different parameters. }
\label{fig:dual-state-rnn}
\end{figure*}

\vspace*{0.05in}
\noindent {\bf DRRN}:  The recurrent structure of DRRN differs only slightly from a ResNet.  In a ResNet, the skip connection comes from the previous residual block, whereas in a DRRN the skip connection always comes from the first unrolled state $s_0$. Figure~\ref{fig:single-state-rnn}(e) shows the equivalent recurrent function for a DRRN with one recursive block (\emph{i.e.} $B=1$) using the definition in the original paper.

\section{Dual-State Recurrent Networks}
Drawing on the connections between state-of-the-art SR models and RNNs, we have investigated new compact RNN architectures for image SR.  Specifically, we propose a dual-state design, which adopts two recurrent states enable use of features from both LR and HR spaces.  The RNN view of our \algname is shown in Figure~\ref{fig:dual-state-rnn}(a) and is introduced as follows. 

\vspace*{0.05in}
\noindent {\bf Dual-state design}:  Unlike single-state models working at the same spatial resolution, \algname incorporates information from both the LR and HR spaces.  Specifically, $s_l$ and $s_h$ in Figure \ref{fig:dual-state-rnn}(a) indicate the LR state and HR state, respectively.  Four colored arrows indicate the transition functions between these two states.  The blue ($f_{\text{lr}}$), orange ($f_{\text{hr}}$) and yellow ($f_{\text{up}}$) links exist in a conventional two-layer RNN, providing information flow from LR to LR, HR to HR, and LR to HR, respectively.  To further enable two-way information flows between $s_l$ and $s_h$, we add the green link, which is inspired by the delayed feedback mechanism of traditional multi-layer RNNs.  Here, it introduces a delayed HR to LR connection.  The overall dynamics of our \algname is given as: 
\begin{equation}
%\small
\begin{aligned}
s_h^t &= f_{\text{up}}(s_l^t) + f_{\text{hr}}(s_h^{t-1}), \text{ and }\\
s_l^t &= f_{\text{input}}(x^t) + f_{\text{lr}}(s_l^{t-1}) + f_{\text{down}}(s_h^{t-1}).
\label{eq:unroll} 
\end{aligned}
\end{equation}
Figure \ref{fig:dual-state-rnn}(b) demonstrates the same concept via an unfolded graph, where the top row represents HR state while the bottom one is LR.   This design choice encourages feature specialization for different resolutions and information sharing across different resolutions.

\vspace*{0.05in}
\noindent {\bf Transition functions}: Our model is characterized by six transition functions.  $f_{\text{up}}$, $f_{\text{down}}$, $f_{\text{lr}}$, and $f_{\text{hr}}$ as illustrated in Figure~\ref{fig:dual-state-rnn}(b).  Specifically, we use the standard residual block for both self-transitions.  A single convolutional layer is used for the down-sampling transition and a single transposed convolutional (or deconvolutional) layer is used for the up-sampling transition. The strides in both inter-state layers are set to be the same as the SR upscaling factor.

\vspace*{0.05in}
\noindent {\bf Unfolding details}: Similarly to unfolding a single-state RNN to obtain a ResNet, for image SR, we let $x^t$ have no contribution to calculating the state transition.  In other words, 
\begin{equation}
f_{\text{input}}(x^t) = 0,  
\end{equation}
for any choice of $x^t$ (\emph{e.g.} choose $x^t =0, ~\forall t$). Furthermore,  we set $s_l^0$ as the output of two convolutional layers with skip connections, which takes the LR input image and transform it into a desired feature space.   In addition, $s_h^0$ is set to zero.  Finally, we use deep supervision for the HR prediction, as discussed below.

\vspace*{0.05in}
\noindent {\bf Deep supervision}:  The unrolled \algname is capable of making a prediction at every time step $t$. Denote 
\begin{equation}
\hat{y}^t = f_{\text{output}}(s_h^t)
\end{equation}
as a prediction at the $t^{th}$ unfolding, where $f_{\text{output}}$ is characterized by a single convolutional layer.  Then, instead of taking the prediction only at the final unfolding $T$, we average all the predictions as 
\begin{equation}
\hat{I_h}=\frac{1}{T}\sum_{t=1}^T \hat{y}^t.
\end{equation}
Thus, every unrolled layer directly connects to the loss layer to facilitate the training of such a very deep network.  Moreover, the model predicts the residual image and minimizes the following mean square error 
\begin{equation}
\label{eq:loss}
\mathcal{L}(\hat{I_h},I_h)= \frac{1}{2} || \hat{I_h} - r_i||^2, 
\end{equation}
where $I_h$ is the group-truth image in HR and $r_i=I_h - bicubic(I_l)$ is the residual map between the ground truth and bicubic upsampled LR image.

\begin{table*}[t]
\centering
\caption{Quantitative evaluation of state-of-the-art SR algorithms: average PSNR/SSIM/IFC for scale factors $\times2$, $\times3$ and $\times4$.  \red{\textbf{Bold red}} text indicates the best and \blue{\underline{underlined blue}} text the second best performance.}
\label{tab:quality} 
%\vspace{1mm}
\begin{tabular}{lccccc}
\hline
\multirow{2}{*}{Algorithm}         & \multirow{2}{*}{Scale} & \textsc{Set5}                                                                  & \textsc{Set14}                                                                 & \textsc{BSDS100}                                                           & \textsc{Urban100}                                                              \\
                                   &                        & PSNR / SSIM / IFC                                                              & PSNR / SSIM / IFC                                                              & PSNR / SSIM / IFC                                                          & PSNR / SSIM / IFC                                                              \\ \hline \hline
Bicubic                            & 2                      & 33.65 / 0.930 / 6.166                                                          & 30.34 / 0.870 / 6.126                                                          & 29.56 / 0.844 / 5.695                                                      & 26.88 / 0.841 / 6.319                                                          \\
A+~\cite{timofte2014a+}            & 2                      & 36.54 / 0.954 / \blue{\underline{8.715}}                                       & 32.40 / 0.906 / 8.201                                                          & 31.22 / 0.887 / 7.464                                                      & 29.23 / 0.894 / 8.440                                                          \\
SRCNN~\cite{dong2014learning}      & 2                      & 36.65 / 0.954 / 8.165                                                          & 32.29 / 0.903 / 7.829                                                          & 31.36 / 0.888 / 7.242                                                      & 29.52 / 0.895 / 8.092                                                          \\
FSRCNN~\cite{dong2016accelerating} & 2                      & 36.99 / 0.955 / 8.200                                                          & 32.73 / 0.909 / 7.843                                                          & 31.51 / 0.891 / 7.180                                                      & 29.87 / 0.901 / 8.131                                                          \\
SelfExSR~\cite{huang2015single}    & 2                      & 36.49 / 0.954 / 8.391                                                          & 32.44 / 0.906 / 8.014                                                          & 31.18 / 0.886 / 7.239                                                      & 29.54 / 0.897 / 8.414                                                          \\
RFL~\cite{schulter2015fast}        & 2                      & 36.55 / 0.954 / 8.006                                                          & 32.36 / 0.905 / 7.684                                                          & 31.16 / 0.885 / 6.930                                                      & 29.13 / 0.891 / 7.840                                                          \\
SCN~\cite{wang2015deep}            & 2                      & 36.52 / 0.953 / 7.358                                                          & 32.42 / 0.904 / 7.085                                                          & 31.24 / 0.884 / 6.500                                                      & 29.50 / 0.896 / 7.324                                                          \\
VDSR~\cite{kim2016accurate}        & 2                      & 37.53 / 0.958 / 8.190                                                          & 32.97 / 0.913 / 7.878                                                          & 31.90 / 0.896 / 7.169                                                      & 30.77 / 0.914 / 8.270                                                          \\
DRCN~\cite{kim2016deeply}          & 2                      & 37.63 / 0.959 / 8.326                                                          & 32.98 / 0.913 / 8.025                                                          & 31.85 / 0.894 / 7.220                                                      & 30.76 / 0.913 / 8.527                                                          \\
LapSRN~\cite{lai2017deep}          & 2                      & 37.52 / 0.959 / \red{\textbf{9.010}}                                           & 33.08 / 0.913 / \red{\textbf{8.505}}                                           & 31.80 / 0.895 / \red{\textbf{7.715}}                                       & 30.41 / 0.910 / \blue{\underline{8.907}}                                       \\
DRRN~\cite{tai2017image}           & 2                      & \red{\bf 37.74} / \blue{\underline{0.959}} / 8.671                             & \red{\bf 33.23} / \red{\bf 0.914} / \blue{\underline{8.320}}                   & \blue{\underline{32.05}} / \blue{\underline{0.897}} / N.A.                 & \red{\bf 31.23} / \red{\bf 0.919} / \red{\bf 8.917}                            \\
\algname                           & 2                      & \blue{\underline{37.66}} / \red{\bf 0.959} / 8.585                             & \blue{\underline{33.15}} / \blue{\underline{0.913}} / 8.169                    & \red{\bf 32.10} / \red{\bf 0.897} / \blue{\underline{7.541}}               & \blue{\underline{30.97}} / \blue{\underline{0.916}} / 8.598                    \\ \hline
Bicubic                            & 3                      & 30.39 / 0.868 / 3.596                                                          & 27.64 / 0.776 / 3.491                                                          & 27.21 / 0.740 / 3.168                                                      & 24.46 / 0.736 / 3.661                                                          \\
A+~\cite{timofte2014a+}            & 3                      & 32.60 / 0.908 / 4.979                                                          & 29.24 / 0.821 / 4.545                                                          & 28.30 / 0.784 / 4.028                                                      & 26.05 / 0.798 / 4.883                                                          \\
SRCNN~\cite{dong2014learning}      & 3                      & 32.76 / 0.908 / 4.682                                                          & 29.41 / 0.823 / 4.373                                                          & 28.41 / 0.787 / 3.879                                                      & 26.24 / 0.800 / 4.630                                                          \\
FSRCNN~\cite{dong2016accelerating} & 3                      & 33.15 / 0.913 / 4.971                                                          & 29.53 / 0.826 / 4.569                                                          & 28.52 / 0.790 / 4.061                                                      & 26.42 / 0.807 / 4.878                                                          \\
SelfExSR~\cite{huang2015single}    & 3                      & 32.63 / 0.908 / 4.911                                                          & 29.33 / 0.823 / 4.505                                                          & 28.29 / 0.785 / 3.922                                                      & 26.45 / 0.809 / 4.988                                                          \\
RFL~\cite{schulter2015fast}        & 3                      & 32.45 / 0.905 / 4.956                                                          & 29.15 / 0.819 / 4.532                                                          & 28.22 / 0.782 / 4.023                                                      & 25.87 / 0.791 / 4.781                                                          \\
SCN~\cite{wang2015deep}            & 3                      & 32.60 / 0.907 / 4.321                                                          & 29.24 / 0.819 / 4.006                                                          & 28.32 / 0.782 / 3.553                                                      & 26.21 / 0.801 / 4.253                                                          \\
VDSR~\cite{kim2016accurate}        & 3                      & 33.66 / 0.921 / 5.088                                                          & 29.77 / 0.834 / 4.606                                                          & \blue{\underline{28.83}} / \blue{\underline{0.798}} / 4.043                & 27.14 / \blue{\underline{0.829}} / 5.045                                       \\
DRCN~\cite{kim2016deeply}          & 3                      & 33.82 / 0.922 / 5.202                                                          & 29.76 / 0.833 / 4.686                                                          & 28.80 / 0.797 / \red{\bf 4.070}                                            & 27.15 / 0.828 / \blue{\underline{5.187}}                                       \\
LapSRN~\cite{lai2017deep}          & 3                      & 33.78 / 0.921 / 5.194                                                          & 29.87 / 0.833 / 4.665                                                          & 28.81 / 0.797 / \blue{\underline{4.057}}                                   & 27.06 / 0.827 / 5.168                                                          \\
DRRN~\cite{tai2017image}           & 3                      & \red{\bf 34.03} / \red{\bf 0.924} / \red{\bf 5.397}                            & \blue{\underline{29.96}} / \blue{\underline{0.835}} / \blue{\underline{4.878}} & \red{\bf 28.95} / \red{\bf 0.800} / N.A.                                   & \red{\bf 27.53} / \red{\bf 0.838} / \red{\bf 5.456}                            \\
\algname                           & 3                      & \blue{\underline{33.88}} / \blue{\underline{0.922}} / \blue{\underline{5.221}} & \red{\bf 30.26} / \red{\bf 0.837} / \red{\bf 4.892}                            & 28.81 / 0.797 / 4.051                                                      & \blue{\underline{27.16}} / 0.828 / 5.172                                       \\ \hline
Bicubic                            & 4                      & 28.42 / 0.810 / 2.337                                                          & 26.10 / 0.704 / 2.246                                                          & 25.96 / 0.669 / 1.993                                                      & 23.15 / 0.659 / 2.386                                                          \\
A+~\cite{timofte2014a+}            & 4                      & 30.30 / 0.859 / 3.260                                                          & 27.43 / 0.752 / 2.961                                                          & 26.82 / 0.710 / 2.564                                                      & 24.34 / 0.720 / 3.218                                                          \\
SRCNN~\cite{dong2014learning}      & 4                      & 30.49 / 0.862 / 2.997                                                          & 27.61 / 0.754 / 2.767                                                          & 26.91 / 0.712 / 2.412                                                      & 24.53 / 0.724 / 2.992                                                          \\
FSRCNN~\cite{dong2016accelerating} & 4                      & 30.71 / 0.865 / 2.994                                                          & 27.70 / 0.756 / 2.723                                                          & 26.97 / 0.714 / 2.370                                                      & 24.61 / 0.727 / 2.916                                                          \\
SelfExSR~\cite{huang2015single}    & 4                      & 30.33 / 0.861 / 3.249                                                          & 27.54 / 0.756 / 2.952                                                          & 26.84 / 0.712 / 2.512                                                      & 24.82 / 0.740 / 3.381                                                          \\
RFL~\cite{schulter2015fast}        & 4                      & 30.15 / 0.853 / 3.135                                                          & 27.33 / 0.748 / 2.853                                                          & 26.75 / 0.707 / 2.455                                                      & 24.20 / 0.711 / 3.000                                                          \\
SCN~\cite{wang2015deep}            & 4                      & 30.39 / 0.862 / 2.911                                                          & 27.48 / 0.751 / 2.651                                                          & 26.87 / 0.710 / 2.309                                                      & 24.52 / 0.725 / 2.861                                                          \\
VDSR~\cite{kim2016accurate}        & 4                      & 31.35 / 0.882 / 3.496                                                          & 28.03 / 0.770 / 3.071                                                          & 27.29 / 0.726 / \blue{\underline{2.627}}                                   & 25.18 / 0.753 / 3.405                                                          \\
DRCN~\cite{kim2016deeply}          & 4                      & 31.53 / 0.884 / 3.502                                                          & 28.04 / 0.770 / 3.066                                                          & 27.24 / 0.724 / 2.587                                                      & 25.14 / 0.752 / 3.412                                                          \\
LapSRN~\cite{lai2017deep}          & 4                      & \blue{\underline{31.54}} / \blue{\underline{0.885}} / \blue{\underline{3.559}} & \blue{\underline{28.19}} / \blue{\underline{0.772}} / \blue{\underline{3.147}} & \blue{\underline{27.32}} / \blue{\underline{0.728}} / \red{\textbf{2.677}} & \blue{\underline{25.21}} / \blue{\underline{0.756}} / \blue{\underline{3.530}} \\
DRRN~\cite{tai2017image}           & 4                      & \red{\textbf{31.68}} / \red{\textbf{0.889}} / \red{\textbf{3.703}}             & \red{\textbf{28.21}} / \red{\textbf{0.772}} / \red{\textbf{3.252}}             & \red{\textbf{27.38}} / \red{\textbf{0.728}} / N.A.                         & \red{\textbf{25.44}} / \red{\textbf{0.764}} / \red{\textbf{3.676}}             \\
\algname                           & 4                      & 31.40 / 0.883 / 3.500                                                          & 28.07 / 0.770 / 3.147                                                          & 27.25 / 0.724 / 2.599                                                      & 25.08 / 0.747 / 3.297                                                          \\ \hline
\end{tabular}
\end{table*}

%%%%%%%%%%%%%%%%%%%%%%%%%%%%%%%%%%%%%%%%%%%%%%%%%%%%%%%%%%%%%%%%%%%%%%
\section{Experiments}
In this section, we first provide implementation details, including both model hyper-parameters and training data augmentation. Then we analyze a number of design choices and their contributions to final performance. Finally, we compare \algname to other state-of-the-art methods on several benchmark datasets. 

\subsection{Datasets}  %erk, the degree stuff, Without the {} the spacing is wrong, and doesn't work in $math$ mode. Yay 40 years of progress in \Latex
To evaluate the proposed \algname algorithm, we train our model using 91 images proposed in \cite{yang2010image} and test on the following datasets: Set5 \cite{bevilacqua2012low}, Set14 \cite{zeyde2010single}, B100 \cite{martin2001database} and Urban100 \cite{huang2015single}. The training data is augmented in a similar way to previous methods \cite{kim2016accurate,tai2017image}, which includes 1) random flipping along the vertical or horizontal axis;  2) random rotation by 90\textdegree{}, 180\textdegree{} or 270\textdegree{}; and  3) random scaling by a factor from [0.5, 0.6, 0.7, 0.8, 0.9, 1]. Tensorflow is used for our full data processing pipeline; the LR training images are generated by the built-in bicubic down-sampling function.  We additionally test our algorithm on 
the DIV2K dataset of the NTIRE SR 2017 challenge \cite{agustsson2017ntire}, where we use the provided training and validation sets with all of the aforementioned data augmentations except random scaling. 

%%%%%%%%%%%%%%%%%%%%%%%%%%%%%%%%%%%%%%%%%%%%%%%%%%%%%%%%%%%%%%%%%%%%%% 
\subsection{Implementation Details}
We use our model to super-resolve only the luminance channel of images, and use bicubic interpolation to upscale the other two color channels, following \cite{kim2016accurate,kim2016deeply,tai2017image}. 
We train independent models for each scale ($\times$2, $\times$3, and $\times$4) with 64 filters on the first input convolutional layer and 128 filters in the rest of the network.  All layers use $3\times3$ convolution filters.  
Due to our dual-state design, the feature maps of $s_l$ and $s_h$ in each time step  have the same spatial dimensions as the LR and HR images, respectively.
We zero-pad the boundaries of feature maps to ensure the spatial size of each feature map is the same as the input size after the convolution is applied.  

% Moreover,  the maximum depth of the unrolled network is controlled by a single hyper-parameter $T$.  In particular, for a \algname with $T$ times unroll, there are at most $5T+3$ convolutional layers from the input to the output, where the multiplier $5$ comes from the two residual blocks plus the up-sampling layer, and the extra 3 are from the input and output layers. 

All the weights in the network are initialized with a uniform distribution using the method proposed in \cite{glorot2010understanding}. We use standard stochastic gradient descent (SGD) with momentum 0.95 as our optimizer to minimize the MSE loss function in Equation (\ref{eq:loss}). We search for the best initial learning rate from $\{0.1,0.03,0.01\}$ and reduce it by a factor of 10 three times during the entire training process.  This learning rate annealing is driven by observing that the loss on the validation set stops decreasing.  Gradient clipping at $0.5$ is adopted during training to prevent the gradient explosion.  We sample image patches with a size of $128\times128$ and use a mini-batch size of $16$ to train our network.

We observe that the recursion defined in Equation (\ref{eq:unroll}) may lead to an exponential increase in the scale of feature values, especially when $T$ is large. In \cite{liao2016bridging},  the authors proposed the use of unshared batch normalization at every unfolding time to resolve this issue.  Batch normalization is not used in our network; we found that normalizing the scale with two scalar parameters was sufficient.  Specifically, we use one unshared PReLU \cite{he2015delving} activation for each recurrent state after every unrolling step.  All other layers have ordinary ReLU as the activation function. 

% \begin{table}[t]
% \caption{\label{tab:self-compare}Performance of \algname (PSNR) on
% the validation set of NTIRE2017 Super-Resolution Challenge, in comparison
% to baseline method and the winning method. }
% \begin{centering}
% \begin{tabular}{ccccccc}
% \hline 
% \multicolumn{1}{c}{} & \multicolumn{3}{c}{Track 1 (bicubic)}\tabularnewline
% \hline 
% \multicolumn{1}{c}{Method} & \multicolumn{1}{c}{x2} & \multicolumn{1}{c}{x3} & \multicolumn{1}{c}{x4} \\
% \hline 
% EDSR+\cite{lim2017enhanced} & 34.93 & 31.13 & 29.09\\
% \hline 
% Single-state Baseline & 34.66 & 30.80 & 28.80 \\
% \hline 
% \algname w/o param sharing & 34.71 &  30.85 &  28.81  \\
% \hline 
% \algname w/o delayed feedback & 34.89 & 30.95	& 28.99 \\
% \hline 
% \algname & \textbf{34.96} & 31.12 & 29.03 \\
% \hline 
% \end{tabular}
% \par\end{centering}
% \end{table}

\begin{table}[t]
\small
\centering
\caption{Quantitative evaluation (in PSNR) of the proposed DSRN, its variants, and other state-of-the-art SR algorithms on track 1 of the NTIRE SR 2017 challenge.  \red{\textbf{Bold red}} text indicates the best and \blue{\underline{underlined blue}} text indicates the second best performance.  The number in $()$ indicates ranking in the challenge.   }
\label{tab:self-compare}
\vspace{1mm}
\begin{tabular}{clccc} 
\hline
\multicolumn{1}{l}{}                    & Method                                         & x2    & x3    & x4    \\ \hline \hline
\multirow{4}{*}{\rotatebox{90}{Ours}}   & Single-state baseline                          & 34.66 & 30.80 & 28.80 \\
                                        & \algname w/o parameter sharing                    & 34.71 & 30.85 & 28.81 \\
                                        & \algname w/o delayed feedback                  & 34.89 & 30.95 & 28.99 \\
                                        & \algname                                       & \red{\bf 34.96} & \blue{\underline{31.12}} & \blue{\underline{29.03}} \\ \hline
\multirow{5}{*}{\rotatebox{90}{Others}} & EDSR+ \cite{lim2017enhanced} (1)               & \blue{\underline{34.93}} & \red{\bf 31.13} & \red{\bf 29.04} \\
                                        & Wang \emph{et al.} \cite{timofte2017ntire} (2) & 34.47 & 30.77 & 28.82 \\
                                        & Bae \emph{et al.} \cite{bae2016beyond} (3)     & 34.66 & 30.83 & 28.83 \\
                                        & SelNet \cite{choi2017deep} (4)                 & 34.29 & 30.52 & 28.55 \\
                                        & BTSRN \cite{fan2017balanced} (5)               & 34.19 & 30.44 & 28.49 \\ \hline
\end{tabular}
\end{table}

\begin{figure}[t]
\begin{centering}
\includegraphics[width=1\columnwidth]{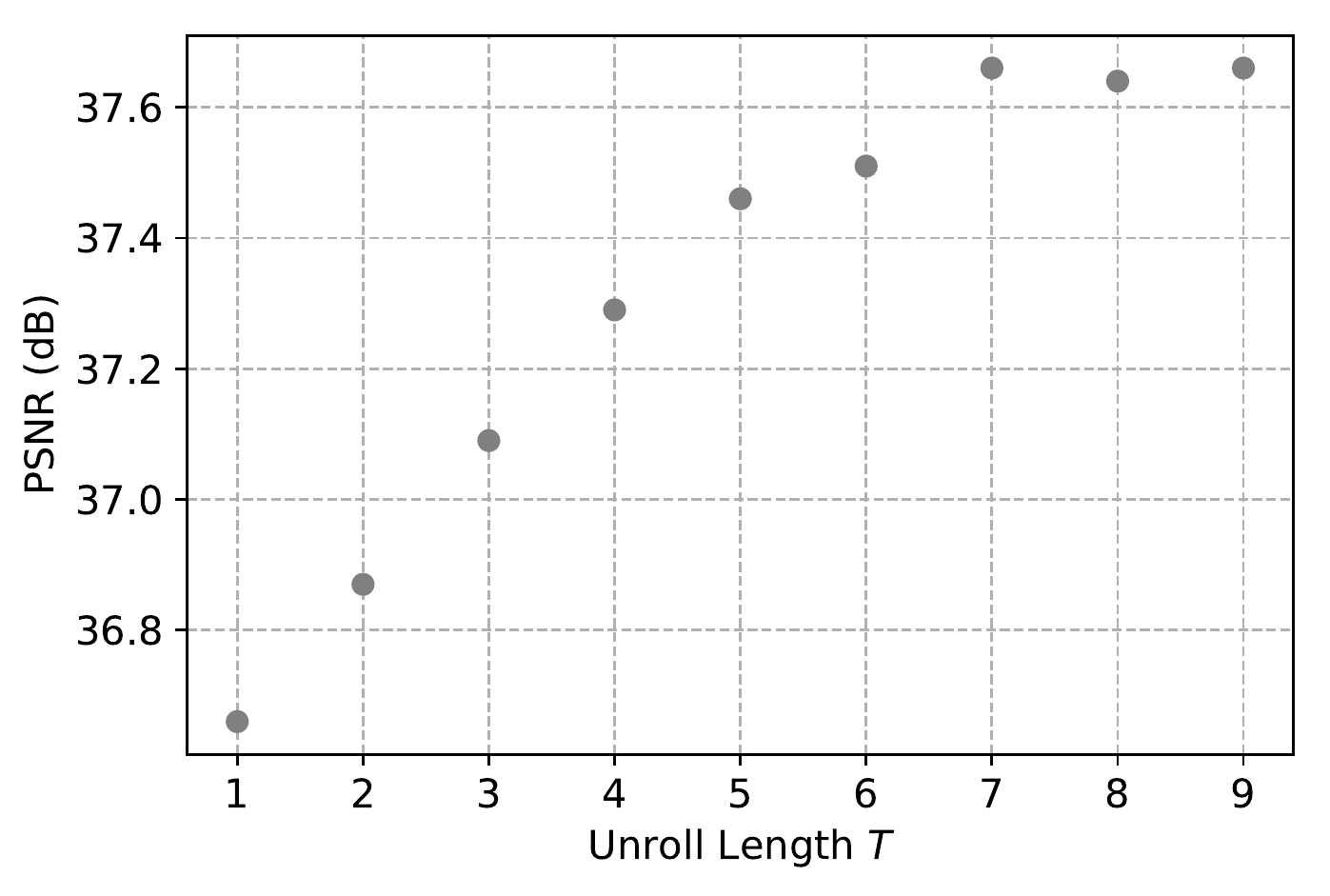}
\par\end{centering}
\caption{Unrolling length v.s. PSNR performance of our \algname with $\times$2 upscaling on Set5 dataset. }
\label{fig:unrollT}
\end{figure}

\begin{figure*}[t!]
\centering

\begin{subfigure}{.24\textwidth}
\includegraphics[width=\textwidth]{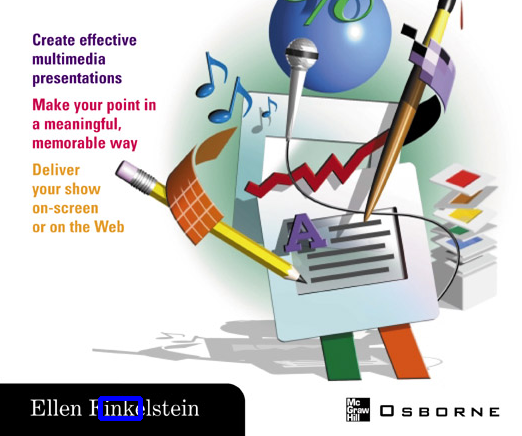}
 \captionsetup{labelformat=empty,font=scriptsize,skip=2pt}
  \caption{Ground truth HR}
\end{subfigure}\hfill
\parbox{.74\textwidth}{
\begin{subfigure}{.24\linewidth}
\includegraphics[width=\textwidth]{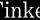}
\captionsetup{labelformat=empty,font=scriptsize,skip=2pt}
\caption{HR}
\end{subfigure}
\begin{subfigure}{.24\linewidth}
\includegraphics[width=\textwidth]{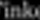}
\captionsetup{labelformat=empty,font=scriptsize,skip=2pt}
\caption{bicubic}
\end{subfigure}
\begin{subfigure}{.24\linewidth}
\includegraphics[width=\textwidth]{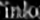}
\captionsetup{labelformat=empty,font=scriptsize,skip=2pt}
\caption{SelfExSR\cite{huang2015single}}
\end{subfigure}
\begin{subfigure}{.24\linewidth}
\includegraphics[width=\textwidth]{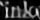}
\captionsetup{labelformat=empty,font=scriptsize,skip=2pt}
\caption{VDSR\cite{kim2016accurate}}
\end{subfigure}\\
\begin{subfigure}{.24\linewidth}
\includegraphics[width=\textwidth]{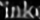}
\captionsetup{labelformat=empty,font=scriptsize,skip=2pt}
\caption{DRCN\cite{kim2016deeply}}
\end{subfigure}
\begin{subfigure}{.24\linewidth}
\includegraphics[width=\textwidth]{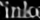}
\captionsetup{labelformat=empty,font=scriptsize,skip=2pt}
\caption{LapSRN\cite{lai2017deep}}
\end{subfigure}
\begin{subfigure}{.24\linewidth}
\includegraphics[width=\textwidth]{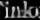}
\captionsetup{labelformat=empty,font=scriptsize,skip=2pt}
\caption{DRRN\cite{tai2017image}}
\end{subfigure}
\begin{subfigure}{.24\linewidth}
\includegraphics[width=\textwidth]{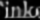}
\captionsetup{labelformat=empty,font=scriptsize,skip=2pt}
\caption{Ours}
\end{subfigure}}

\begin{subfigure}{.24\textwidth}
\includegraphics[width=\textwidth]{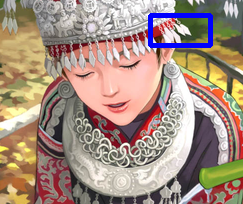}
 \captionsetup{labelformat=empty,font=scriptsize,skip=2pt}
  \caption{Ground truth HR}
\end{subfigure}\hfill
\parbox{.74\textwidth}{
\begin{subfigure}{.24\linewidth}
\includegraphics[width=\textwidth]{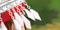}
\captionsetup{labelformat=empty,font=scriptsize,skip=2pt}
\caption{HR}
\end{subfigure}
\begin{subfigure}{.24\linewidth}
\includegraphics[width=\textwidth]{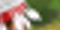}
\captionsetup{labelformat=empty,font=scriptsize,skip=2pt}
\caption{bicubic}
\end{subfigure}
\begin{subfigure}{.24\linewidth}
\includegraphics[width=\textwidth]{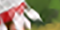}
\captionsetup{labelformat=empty,font=scriptsize,skip=2pt}
\caption{SelfExSR\cite{huang2015single}}
\end{subfigure}
\begin{subfigure}{.24\linewidth}
\includegraphics[width=\textwidth]{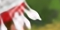}
\captionsetup{labelformat=empty,font=scriptsize,skip=2pt}
\caption{VDSR\cite{kim2016accurate}}
\end{subfigure}\\
\begin{subfigure}{.24\linewidth}
\includegraphics[width=\textwidth]{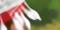}
\captionsetup{labelformat=empty,font=scriptsize,skip=2pt}
\caption{DRCN\cite{kim2016deeply}}
\end{subfigure}
\begin{subfigure}{.24\linewidth}
\includegraphics[width=\textwidth]{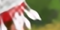}
\captionsetup{labelformat=empty,font=scriptsize,skip=2pt}
\caption{LapSRN\cite{lai2017deep}}
\end{subfigure}
\begin{subfigure}{.24\linewidth}
\includegraphics[width=\textwidth]{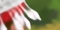}
\captionsetup{labelformat=empty,font=scriptsize,skip=2pt}
\caption{DRRN\cite{tai2017image}}
\end{subfigure}
\begin{subfigure}{.24\linewidth}
\includegraphics[width=\textwidth]{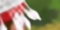}
\captionsetup{labelformat=empty,font=scriptsize,skip=2pt}
\caption{Ours}
\end{subfigure}}

\begin{subfigure}{.24\textwidth}
\includegraphics[width=\textwidth]{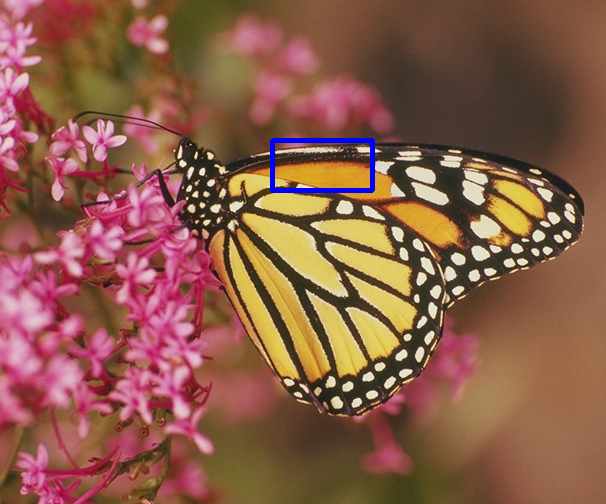}
 \captionsetup{labelformat=empty,font=scriptsize,skip=2pt}
  \caption{Ground truth HR}
\end{subfigure}\hfill
\parbox{.74\textwidth}{
\begin{subfigure}{.24\linewidth}
\includegraphics[width=\textwidth]{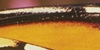}
\captionsetup{labelformat=empty,font=scriptsize,skip=2pt}
\caption{HR}
\end{subfigure}
\begin{subfigure}{.24\linewidth}
\includegraphics[width=\textwidth]{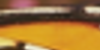}
\captionsetup{labelformat=empty,font=scriptsize,skip=2pt}
\caption{bicubic}
\end{subfigure}
\begin{subfigure}{.24\linewidth}
\includegraphics[width=\textwidth]{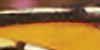}
\captionsetup{labelformat=empty,font=scriptsize,skip=2pt}
\caption{SelfExSR\cite{huang2015single}}
\end{subfigure}
\begin{subfigure}{.24\linewidth}
\includegraphics[width=\textwidth]{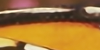}
\captionsetup{labelformat=empty,font=scriptsize,skip=2pt}
\caption{VDSR\cite{kim2016accurate}}
\end{subfigure}\\
\begin{subfigure}{.24\linewidth}
\includegraphics[width=\textwidth]{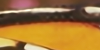}
\captionsetup{labelformat=empty,font=scriptsize,skip=2pt}
\caption{DRCN\cite{kim2016deeply}}
\end{subfigure}
\begin{subfigure}{.24\linewidth}
\includegraphics[width=\textwidth]{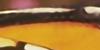}
\captionsetup{labelformat=empty,font=scriptsize,skip=2pt}
\caption{LapSRN\cite{lai2017deep}}
\end{subfigure}
\begin{subfigure}{.24\linewidth}
\includegraphics[width=\textwidth]{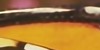}
\captionsetup{labelformat=empty,font=scriptsize,skip=2pt}
\caption{DRRN\cite{tai2017image}}
\end{subfigure}
\begin{subfigure}{.24\linewidth}
\includegraphics[width=\textwidth]{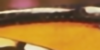}
\captionsetup{labelformat=empty,font=scriptsize,skip=2pt}
\caption{Ours}
\end{subfigure}}

\begin{subfigure}{.24\textwidth}
\includegraphics[width=\textwidth]{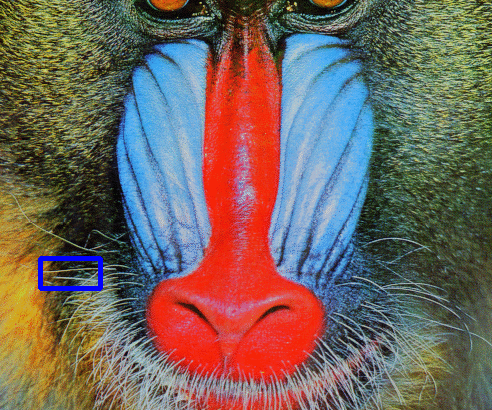}
 \captionsetup{labelformat=empty,font=scriptsize,skip=2pt}
  \caption{Ground truth HR}
\end{subfigure}\hfill
\parbox{.74\textwidth}{
\begin{subfigure}{.24\linewidth}
\includegraphics[width=\textwidth]{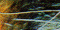}
\captionsetup{labelformat=empty,font=scriptsize,skip=2pt}
\caption{HR}
\end{subfigure}
\begin{subfigure}{.24\linewidth}
\includegraphics[width=\textwidth]{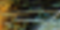}
\captionsetup{labelformat=empty,font=scriptsize,skip=2pt}
\caption{bicubic}
\end{subfigure}
\begin{subfigure}{.24\linewidth}
\includegraphics[width=\textwidth]{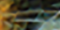}
\captionsetup{labelformat=empty,font=scriptsize,skip=2pt}
\caption{SelfExSR\cite{huang2015single}}
\end{subfigure}
\begin{subfigure}{.24\linewidth}
\includegraphics[width=\textwidth]{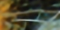}
\captionsetup{labelformat=empty,font=scriptsize,skip=2pt}
\caption{VDSR\cite{kim2016accurate}}
\end{subfigure}\\
\begin{subfigure}{.24\linewidth}
\includegraphics[width=\textwidth]{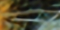}
\captionsetup{labelformat=empty,font=scriptsize,skip=2pt}
\caption{DRCN\cite{kim2016deeply}}
\end{subfigure}
\begin{subfigure}{.24\linewidth}
\includegraphics[width=\textwidth]{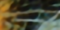}
\captionsetup{labelformat=empty,font=scriptsize,skip=2pt}
\caption{LapSRN\cite{lai2017deep}}
\end{subfigure}
\begin{subfigure}{.24\linewidth}
\includegraphics[width=\textwidth]{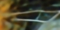}
\captionsetup{labelformat=empty,font=scriptsize,skip=2pt}
\caption{DRRN\cite{tai2017image}}
\end{subfigure}
\begin{subfigure}{.24\linewidth}
\includegraphics[width=\textwidth]{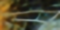}
\captionsetup{labelformat=empty,font=scriptsize,skip=2pt}
\caption{Ours}
\end{subfigure}}
\captionsetup{skip=60pt}
\caption{Qualitative Comparison on Set 14 with $\times 3$ upscaling. From top to bottom: 
 1) the image "ppt3", \algname reconstructs sharp text with less artifacts. 2) the image "comic". 3) the image "monarch", \algname finds less blurry dots along the edge of the wing. 4) the image "baboon".}
\label{fig:qualitative}
\end{figure*}

%%%%%%%%%%%%%%%%%%%%%%%%%%%%%%%%%%%%%%%%%%%%%%%%%%%%%%%%%%%%%%%%%%%%%%
\subsection{Model Analysis}
In this section, we analyze our proposed model in the following respects:  

\begin{figure*}[tb]
\begin{centering}
\includegraphics[width=1\textwidth]{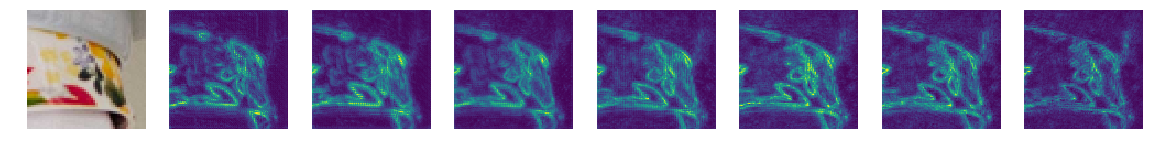}
\par\end{centering}
\caption{Feature visualization: input image patch and the energy maps of the output at HR states (7 unrolled timestamps in total). }
\label{fig:vis}
\end{figure*}

%\vspace*{0.05in}
\noindent {\bf Unrolling length}: The unrolling length $T$ changes the maximum effective depth of the unrolled network.  In particular, for a \algname with $T$ times unrolling, the maximum number of convolution layers between input and output of the network is  $2T+4$.  
The multiplier $2$ comes from the two layers in a residual block, while the extra 4 is from the auxiliary input and output layers.
However,  the number of model parameters remains independent of the length of unrolling.   Essentially, $T$ controls the trade-off between model capacity and computation cost. We study the influence of $T$ by training the model with different unrolling lengths.   The empirical results are shown in Figure~\ref{fig:unrollT}.  The test performance increases when the number of unfolding steps increases, but the benefit seems to diminish after $T=7$.  Unless otherwise mentioned, we use $T=7$ for all our models.  It is worth mentioning that we also experimented with stochastic depth \cite{huang2016deep} by randomly sampling $T$ during training, but we observed no improvement in validation accuracy.

%\vspace*{0.05in}
\noindent {\bf Parameter sharing}:   We empirically find parameter sharing to be crucial for training a deep recursive model.  As shown in Table~\ref{tab:self-compare}, the same model with untied weights performs much more poorly than its weight-sharing counterpart.  
Specifically, we observe around 0.2dB performance drop across all three upscaling scales when changing from shared weights to untied weights.
We speculate that the model with untied weights suffers a larger risk of model over-fitting and much slower training convergence, both of which diminish the model's restoration accuracy.
%Note that the comparison might not be conclusive because that our attempt to train the unshared model always ends up with significantly slower convergence speed. Due to the time limit, we cannot verify if the unshared model will eventually converge to a better solution. 

\vspace*{0.05in}
\noindent {\bf Dual-state and delayed feedback}: We compare our \algname with two baselines under the same unrolling time steps to understand how each module of our model contributes to the final performance: 1) a single-state RNN unrolled ResNet; and 2) a dual-state RNN without delayed feedback connections.   
The quantitative comparison on the NTIRE SR 2017 challenge is shown in Table~\ref{tab:self-compare}.   
Comparing the single-state baseline and the \algname without feedback, it is clear that considering information from both LR and HR spaces as two separated states provides performance gains.   In addition, comparing our models with and without feedback, we realize that incorporating such an information flow from HR space back to LR space consistently improves performance on all three different scales.   In all, both the dual-state and delayed feedback designs are beneficial to our model.

%\vspace*{0.05in}
\noindent {\bf State visualization} Since DSRN has independent scaling parameters on each unrolled state, the model implicitly learns a weighted-average of all the unrolled states for the final prediction. Empirically we observe that this strategy performs better than output from the last state only. To demonstrate how the network aggregates different unrolled states, we show feature response maps at different unrolling steps in Figure~\ref{fig:parampsnr}, demonstrating that the network distributes slightly different features to each unrolled state.

%make the model more competitive.  

%%%%%%%%%%%%%%%%%%%%%%%%%%%%%%%%%%%%%%%%%%%%%%%%%%%%%%%%%%%%%%%%%%%%%%

\subsection{Comparison with the State-of-the-Art}

We provide results of evaluation of our model on several public benchmark datasets in Table \ref{tab:quality}, with three commonly-used evaluation metrics: Peak Signal-to-Noise Ratio (PSNR), Structural SIMilarity (SSIM) \cite{wang2004image} and the Information Fidelity Criterion (IFC) \cite{sheikh2005information}.  Specifically, we perform a comprehensive comparison between our method and 10 other existing SR algorithms, including both deep learning and non-deep-learning based methods.  Note that many recent deep learning based competitors, including VDSR, LapSRN and DRRN,  use 291 training samples with the additional 200 from the training set of Berkeley Segmentation Dataset \cite{arbelaez2011contour}, while our model was trained on only the 91 images.  Still, our \algname method achieves competitive performance across all datasets and scales.  It achieves particularly strong performance in the $\times2$ and $\times3$ settings. 

In addition, we report quantitative evaluations on the recently developed DIV2K dataset and comparisons with top-ranking algorithms in Table~\ref{tab:self-compare}. Our method achieves competitive performance with the best algorithm, EDSR+\cite{lim2017enhanced}, and outperforms all the other algorithms by a large margin, which demonstrates the effectiveness of our proposed dual-state recurrent structure.

To further analyze the proposed \algname against other state-of-the-art SR approaches in a qualitative manner, in Figure \ref{fig:qualitative} we present several visual examples of super-resolved images  on Set14 with $x3$ upscaling 
%as the visual comparison. 
among different SR approaches. For these competing methods, we use SR results publicly released by the authors.   As shown in Figure~\ref{fig:qualitative}, our method can construct sharp and detailed structures and is less prone to generating spurious artifacts. 

\begin{figure}[t!]
\begin{centering}
\includegraphics[width=1\columnwidth]{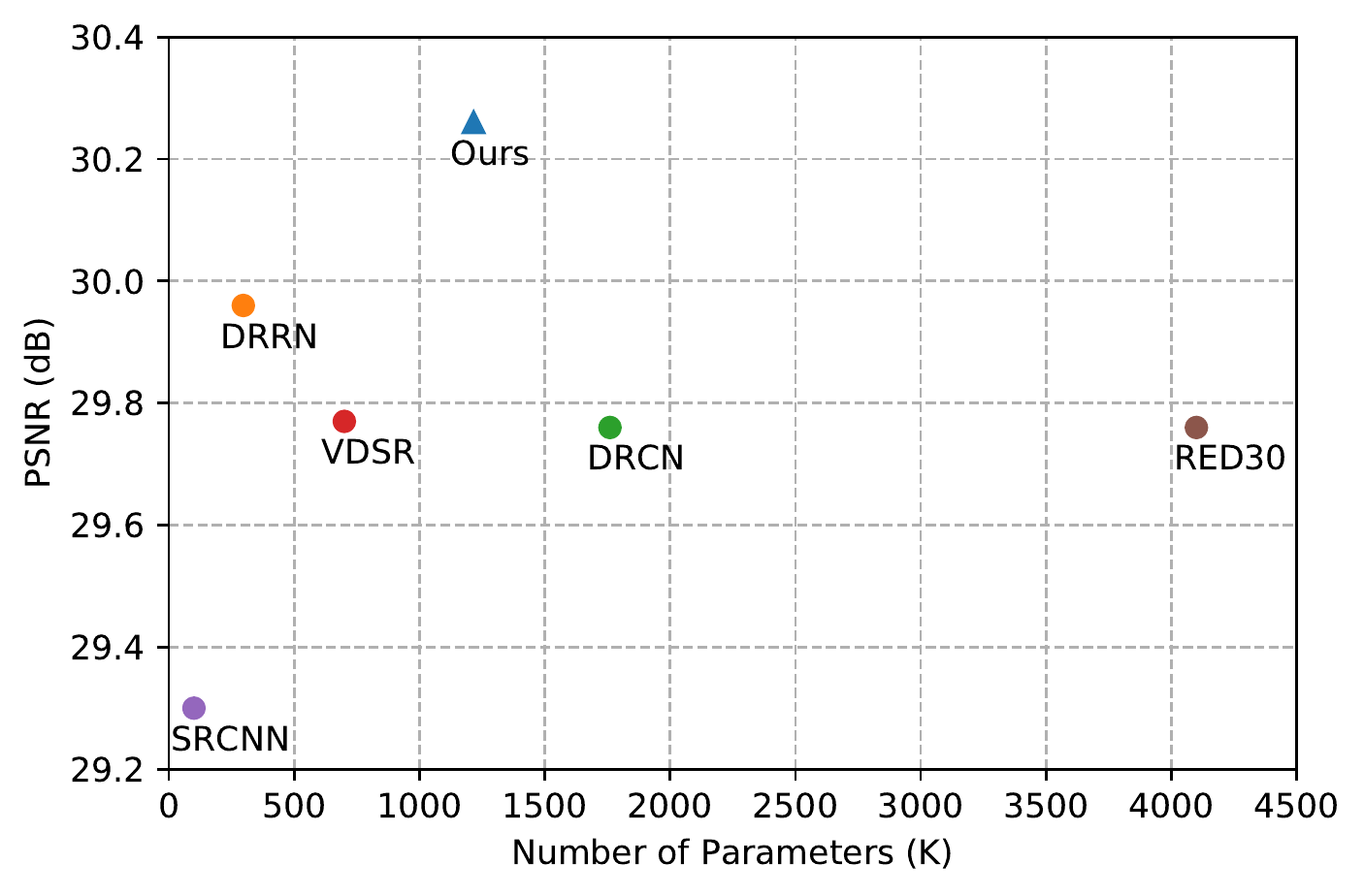}
\par\end{centering}
\caption{Comparison of the PSNR and the model size of  recent SR methods for  $\times 3$ upscaling on Set 14.}
\label{fig:parampsnr}
\end{figure}

Furthermore, the proposed \algname benefits from inherent parameter sharing and therefore obtains higher parameter efficiency compared to other methods. In Figure~\ref{fig:parampsnr}, we illustrate the parameters-to-PSNR relationship of our model and several state-of-the-art methods, including SRCNN, VDSR, DRCN, DRRN and RED30 \cite{mao2016image}. Our method represents a favorable trade-off between model size and SR performance, and has modest inference time. The \algname takes 0.4s on the x4 task with a 288x288 output image size, on an NVIDIA Titan X GPU. 

\section{Conclusion}
In this work, we have provided a unique formulation that expresses many state-of-the-art SR models as a finite unfolding of a single-state RNN with various recurrent functions.  Based on this, we extend existing methods by considering a dual-state design; the two hidden states of our proposed \algname operate at different spatial resolutions. One captures the LR information while the other one targets the HR domains.  To ensure two-way communication between states, we integrate a delayed feedback mechanism.  Thus, the predicted features from both LR and HR states can be exploited jointly for final predictions.  Extensive experiments on benchmark datasets have demonstrated that the proposed \algname performs favorably against state-of-the-art SR models in terms of both efficiency and accuracy. For the future work, we will explore use of our proposed DSRN to capture temporal dependencies for video SR~\cite{liu2017robust}. 

%%%%%%%%%%%%%%%%%%%%%%%%%%%%%%%%%%%%%%%%%%%%%%%%%%%%%%%%%%%%%%%%%%%%%%%%%%%%%%%
%\section*{Acknowledgement}
%This section will be added upon the acceptance of this paper.  

% ============== version 1 ====================
% In this work, we explored a number of novel neural network structure constructed from unrolled HO-er of novel neural network structure constructed from unrolled HO-RNN, on the task of single image super-resolution. We found that: 1) unrolled HO-RNN provide a convenient and intuitive way for neural network structure design. 2) A specific HO-RNN model works well for single image super-resolution. 3) The model naturally reuse layers and is more parameter efficient comparing with other methods. We believe the methodology can be applied to other computer vision tasks such as object detection as well, which is our future work. 

%%%%%%%%%%%%%%%%%%%%%%%%%%%%%%%%%%%%%%%%%%%%%%%%%%%%%%%%%%%%%%%%%%%%%%%%%%%%%%%
% \newpage
% \newpage
% \newpage
% \newpage
\bibliographystyle{ieee}
\bibliography{egbib}

\end{document}